\documentclass[runningheads,table]{llncs}
\usepackage[T1]{fontenc}
\usepackage{graphicx}
\usepackage{color}
\usepackage{microtype}
\usepackage{graphicx}
\usepackage{booktabs} % for professional tables

\usepackage{amsmath}
\usepackage{amssymb}
\usepackage{mathtools}
\usepackage{bbm}

\usepackage[textsize=tiny]{todonotes}

\graphicspath{ {./imagesP2/} }
\usepackage{caption}
\usepackage{subcaption}
\usepackage{float}
\usepackage{booktabs}
\usepackage{soul}

\providecommand{\manisha}[1]{\textcolor{blue}{MS: {#1}} }

\begin{document}
\title{EMOTE: An Explainable architecture for Modelling the Other Through Empathy}
\titlerunning{EMOTE}
\author{Manisha Senadeera \and
Thommen Karimpanal George \and
Sunil Gupta \and
Stephan Jacobs \and
Santu Rana}
%
%\author{Manisha Senadeera}

\authorrunning{M. Senadeera et al.}
%\institute{Princeton University, Princeton NJ 08544, USA
%\url{http://www.springer.com/gp/computer-science/lncs}

\institute{Applied Artificial Intelligence Institute, Deakin University \email{manisha.senadeera@deakin.edu.au}}

\maketitle              % typeset the header of the contribution
\begin{abstract}

We can usually assume others have goals analogous to our own. This assumption can also, at times, be applied to multi-agent games - e.g. Agent 1's attraction to green pellets is analogous to Agent 2's attraction to red pellets. This ``analogy'' assumption is tied closely to the cognitive process known as empathy. Inspired by empathy, we design a simple and explainable architecture to model another agent's action-value function. This involves learning an \emph{Imagination Network} to transform the other agent's observed state in order to produce a human-interpretable \emph{empathetic state} which, when presented to the learning agent, produces behaviours that mimic the other agent. Our approach is applicable to multi-agent scenarios consisting of a single learning agent and other (independent) agents acting according to fixed policies. This architecture is particularly beneficial for (but not limited to) algorithms using a composite value or reward function. We show our method produces better performance in multi-agent games, where it robustly estimates the other's model in different environment configurations. Additionally, we show that the empathetic states are human interpretable, and thus verifiable.

\end{abstract}
%
%
%

% MAIN BODY
\section{Introduction}
Getting along with everyone isn't easy, but trying to understand them is an essential first step to a smooth and pleasant encounter. 
%but unfortunately life contains plenty of interactions with difficult and unpleasant people\thommen{I feel like this first sentence is not directly relevant. Why not simply say ` To effectively deal with everyday interactions, understanding the other is a an essential quality.'} \stephan{The point is perhaps that we particularly need to want to ``understand the other'' when we have very little in common with them. So maybe 'Getting along with everyone isn't easy, but unfortunately life contains plenty of interactions with people we cannot relate to'}. In an attempt to get through some encounters unscathed, understanding the other is a productive step. 
To support this, most of us possess the cognitive ability known as empathy. Hoffman \cite{hoffman1996empathy} defines empathy as \textit{``any process where the attended perception of the object’s state generates a state in the subject that is more applicable to the object’s state or situation than to the subject’s own prior state or situation."} From this definition, empathy can be interpreted as a process which helps us understand the feelings and goals of another, by using ourselves as a point of reference. \textit{How would we feel if we were in a similar situation? What objects or goals, though not exactly the same, do we feel similarly towards?} An implicit assumption is the belief that who we are observing is analogous or similar to us, even if our goals or preferences are different. %\st{Imagine seeing a dog in pain from an injury. Of course, there are many differences between us and it, but we can empathise with it's pain because we can imagine how it might feel, based on how we would feel in the same situation. Now simply putting ourselves in another's shoes is not always enough to empathise 
%(not only because your feet might not fit, or because dogs don't wear shoes)\thommen{The joke might be distracting, maybe even confusing for some readers who need not share the same sense of humour/cultural background. Also, I think it doesn't add much to the main point - it is currently an aside.} \stephan{``Dogs don't wear shoes'' is an example of a difference in experience that makes it difficult to empathise, I think.}, 
%because the preferences we have aren't always exactly the same.} 
As a simple example: I love chocolate but dislike liquorice. If I were to see someone eating and enjoying liquorice, I would not immediately relate to how they feel. However, if I \textit{imagined} it was instead chocolate they are consuming, I could understand their joy and subsequently infer the levels of enjoyment as being similar to mine when eating chocolate. In this situation, the chocolate and liquorice are analogous features.
%\thommen{is it that we imagine they are eating chocolate, or that we imagine their enjoyment being similar to ours when we eat chocolate?} \stephan{Given the mechanism of the framework in the paper I believe the intent here is to imagine them eating chocolate: ``How does the underlying \emph{state} change in order to evoke the same reward as a consequence?''}\manisha{I've changed the following sentence for more clarity.} 

%\st{Empathy between humans is not easy, but it pales in comparison with empathy between artificial agents. Trickier still: between a human and an agent!}
%\thommen{The last few sentences are a bit informal. It is okay, but alternating tones from less formal to formal could be somewhat jarring to readers.} 
With the increasing prevalence of robots, virtual assistants, etc. in regular life, artificially intelligent systems in shared environments will benefit from modelling and understanding other agents or humans around them. % \st{especially if we hope for safe and trust-filled engagements}. This problem falls within the space of multi-agent reinforcement learning (MARL) \cite{hu1998multiagent}. 
%\st{In multi-agent scenarios, it is common to train all agents simultaneously.}\thommen{I think this line can be removed} 
A subset of this space involves a single learning agent coexisting with, and learning to model, one or more other agents who behave under fixed policies (e.g. a human with set preferences or a pre-trained robot) \cite{AIJ18-Albrecht}. 
%\st{One can imagine this scenario to be like that of a new robot entering a room with a human (who already has a set of preferences and behaviours) or a pre-trained robot, where the new robot is trying to learn how to achieve its own goals in the presence of the other.}

%Though this problem can be thought of as training a single RL agent with all other agents considered part of the environment, it can instead be beneficial to explicitly model the other. \thommen{I found this paper: https://www.cs.utexas.edu/~pstone/Papers/bib2html/b2hd-AIJ18-Albrecht.html It seems like this is already an established sub-field in multiagent learning, so I'm wondering if we should just say that it falls under a subset of multiagent learning where one agent models other agents.}%Humans are, after all, a bit complicated.

To incorporate empathetic modelling behaviours 
%\thommen{to incorporate empathetic behaviours or to enable agents to appropriately model the interests of others in such scenarios?} 
in agents being trained in such scenarios,
%\thommen{The previous para ended with just describing the situation of a human and pretrained robots. Now when we start this para with `Towards this pursuit', its not clear what we mean. Thats why I suggested `To incorporate empathetic behaviours to agents being trained in such scenarios, we present...' or something like that}
%\thommen{when you say `for this', it is not clear what you are referring to. You could add something like `To incorporate empathetic behaviours to agents being trained in such scenarios, we present...'}, 
we present EMOTE - an \emph{Explainable architecture for Modelling the Other Through Empathy}. Drawing inspiration from empathy, this architecture is designed to allow a learning agent to reference its own action-value function to model another agent's action-value function, leading to a more stable and robust representation of the other. Crucially, leveraging the learning agent's own functions enables the other agent's model to be \emph{human-interpretable}, allowing interrogation of the inferences made by the learning agent
about the other agent's goals (in the form of a reward function). %\st{to model the other agent's action-value function.} 
We consider settings where analogous agents share a common environment - specifically, a single learning agent aims to model the other `independent' agents who are %\st{by referencing its own action-value function}. \st{The independent agents are} 
pre-trained and act according to fixed policies, the reward functions of which the learning agent is not privy to.
%, but aims to infer.

EMOTE consists of a two stage neural network architecture. The first, called the \emph{Imagination Network}, imagines an empathetic representation of the independent agent's state. This \textit{empathetic state} can be understood to be the perception of the independent agent's state, from the perspective of the learning agent (e.g. liquorice reimagined as chocolate). This empathetic state is fed into a second network, a copy of the learning agent's own action-value function, to observe what values the learning agent would have associated with this empathetic state.

The benefits of EMOTE are three fold. Firstly, by referencing its own action-value function, the reward and action-value estimates made by the learning agent about the independent agent are more robust which we demonstrate in different environment settings (adversarial/assistive) and configurations (layouts).
%\thommen{I think its robustness and stability in terms of estimating the rewards/action-values? Is it referring to changes in the env layout?} I  changed slightly.} 
Secondly, explaining the inferences made by the learning agent about the independent agent's behaviour is possible through the generated empathetic state which is human-interpretable. A user can tap into this empathetic state and compare it against the original state of the independent agent, observing which features remain the same (both agents view and react similarly to), and which have changed (e.g. the liquorice being reimagined as chocolate), offering a useful tool for interpretable verification of the independent agent's action values.
Lastly, funnelling the empathetic state through the learning agent's own action-value function has a constraining effect
%\thommen{why is this different from the first one? Cant they be combined?}\manisha{It is different cos this last one is about ensuring the scales are similar. The first one is about getting the same results even if the environment layout changes.} 
such that the action-values and corresponding inferred rewards, lie in a similar scale to that of the learning agent. Often in multi-agent games, knowledge of the independent agent is used by the learning agent to guide its own behaviour. A common approach to achieve this is by constructing a composite reward function, value function or policy \cite{Noothigattu2019,alamdari2021considerate}. In such approaches, ensuring similarity in value scales of the functions being combined is important for stable performance. Fortunately our architecture naturally ensures that the range of the inferred action-value and reward functions will be comparable to those of the learning agent, obviating the need for scaling to construct the composite function. One may also view this characteristic through the lens of inverse reinforcement learning (IRL), where the goal is to map observed behaviours to a reward function. A key challenge here is that there exists a large space of reward functions which could possibly correspond to a given arbitrary behaviour. By inferring the reward function of another agent through the use of the learning agent's own action-value function, our architecture offers an elegant solution to naturally narrow down the space of candidate reward functions, while also enabling analogous features between the agents to be mapped to similar reward ranges.
%\st{An estimate of the independent agent's reward function can thus be derived from the EMOTE action-value function, which will also be of a comparable range to the learning agent's reward function. This characteristic is particularly useful in scenarios involving composite action-value functions or reward functions where the range of the functions being combined is important for stable performance, as is often the case for many multiagent algorithms}. 

We demonstrate our proposal by integrating our EMOTE architecture in the work by Senadeera et.al  \cite{SympathyPaper}. %\st{leverage previous work by} %\st{and adapt it to integrate our EMOTE architecture.} 
Their framework focused on
building considerate learning agents by flexibly combining the learning and independent agents' reward and action-value functions. 
%\st{, thereby providing a platform to illustrate \manisha{two examples of EMOTE's second benefit.}}
%\manisha{ thus enabling the illustration of EMOTE's first benefit. Applied to assistive and adversarial multiagent scenarios,} 
Applied to assistive and adversarial multiagent scenarios, we show that EMOTE produces more stable models of the independent agent that are robust to various settings and layouts.
%\manisha{performance meeting the state-of-the-art.}\thommen{I think its better not to mention SoA because later on, we argue that most of the SoA methods are not applicable}\manisha{What should I say then instead? ..leading to good performance?}\thommen{maybe something like '..leading to consistently good performances in a variety of environment settings, including those involving assistive as well as adversarial agents.' }.
%\thommen{in assistive and adversarial multiagent scenarios?}. 
Further, we observe the Imagination Network is capable of recovering interpretable empathetic states (i.e., finding analogous features between the agents). The primary contribution of this work is an architecture that produces a stable model of the independent agent's action-value function. As a result of our design, the model permits human-interpretability of the inferences made by the learning agent about the independent agent's value function whilst ensuring these inference values are in a similar scale to that of the learning agent. Our method does not aim to outperform current state-of-the-art, rather to produce robust and stable inferences about the independent agent's action-value and reward function. We assume that if inferred correctly, performance result will at least matches the state-of-the-art.

\section{Related Literature}

\paragraph{Modelling the other:} Modelling other agents is a key component of multi-agent reinforcement learning (MARL). There exists a vast body of work ranging from inferring the other's policy \cite{foerster:aamas18,wen2019probabilistic,hu2020other,shu2018m}, goals and beliefs \cite{Raileanu2018ModelingOU,moreno2021neural} and value functions \cite{zhao2022mcmarl,pmlr-v48-he16}. These works predominantly assume that all agents are being trained concurrently which differs from our intended setting where only a single agent is trained. Modelling agents who behave according to a fixed pre-trained policy can be tackled with works in Theory of Mind (ToM) \cite{pmlr-v80-rabinowitz18a} and Inverse Reinforcement Learning (IRL) \cite{ng2000algorithms} literature. A small subset of MARL combines these two problems to create environments in which a learning agent (to be trained) coexists with and tries to model a pre-trained (independent) agent \cite{papoudakis2021agent,SympathyPaper}, in line with the problem setting of our work. However, such works generally do not produce interpretable models, and produce arbitrarily scaled action-value estimates, which impedes the accurate inference of agent behaviours.% \manisha{added in previous line}
%\thommen{Can we say something like `However, such works generally do not produce interpretable models, and produce arbitrarily scaled action-value estimates, which impedes the accurate inference of agent behaviours.'} %Our work targets this subset.
%One such paper is that by  in which the learning agent is assumed to only have access to the other agent's trajectories during training. As such, to incorporate the other's behaviour into the learning agent's policy, a latent representation is extracted and trained on. \thommen{What is this paper's relation to the current work?}

\paragraph{Modelling based on oneself:} A selection of works model the other agent based on their own model. Using ToM, \cite{Raileanu2018ModelingOU} trains the learning agent on all possible goals during training and uses this information to infer the hidden goal of the other agent based on its behaviour. This method differs from our setting as it is constrained to games that have set goals which can be experienced by the learner. Inspired by empathy as well, \cite{TowardsEmpathicDQN} proposes 
%has a learning agent that shares an environment with a single independent agent. In order 
imposing the learning agent's own value function directly on the independent agent, using this to infer the other's intent. A limitation is that imposing the same value function on the independent agent assumes this agent has the same values as the learner. Our work eschews this assumption, allowing for different and even opposing intentions. 

\paragraph{Composite value and reward functions:} In multi-agent scenarios with composite reward or value functions (e.g. summation of two or more reward or value estimates), it is important they are scaled appropriately to ensure stable behaviours. %\st{ensuring comparability}\thommen{it might be unclear what we mean by comparability. How about `When summating two or more reward or value estimates, to achieve stable behaviours, it is important to ensure they are scaled appropriately.'} \st{is important to produce stable behaviours.} 
 Approaches such as VDN \cite{VDN} and QMIX \cite{rashid2018qmix} combine separate agent value functions to conduct centralised training, thus obviating the need for such scaling. However, this does not allow for independent agents with pre-trained policies. %When all agents are being trained, these value functions do not need any further scaling or adjustment prior to combining them together. 
 More closely related, \cite{alamdari2021considerate} builds a joint function of learning agent and independent agent rewards (whose rewards are already known). A similar joint reward function is built by \cite{SympathyPaper} however they use IRL to to infer the rewards of the independent agents.
%\st{More closely related algorithms include that by \cite{alamdari2021considerate}, where a joint function of the learning agent's reward together with the reward of all independent agents is constructed to train the learning agent. No scaling was applied, as it was assumed that a distribution over reward functions for the independent agent was known. \cite{SympathyPaper} also trains a learning agent on a similar joint reward function, with the independent agent's reward function being inferred via IRL.}
As a result of the space of potential rewards that can emerge through IRL \cite{PolicyInvariance}, the paper mitigates the issues of a misalignment by scaling the independent agent's functions by a constant (the ratio of the $l1$ norms of the learning agent's reward vector and the IRL inferred rewards of the independent agent). This method was also applied by \cite{Noothigattu2019}. %\st{, where an ethical agent was trained using a multi-arm bandit reward function (using a combination of rewards).} 
This simple $l1$ norm based normalisation may fail in many complex scenarios, and additionally, is constrained to problems that only sum two reward or action-value functions, motivating need for alternatives. %Our proposed work aims to address this issue, along with other  \manisha{one of which our work proposes.}

\section{Methodology}\label{sec:method}

\subsection{Problem Setting}\label{sec:problem}
\iffalse
\begin{figure}[h]
  \centering
  \includegraphics[width=\linewidth]{Project2_framework}
  \caption{Sympathy Framework with proposed empathy-based architecture for independent agent's value function ($\hat{Q}_{indep}$). $\hat{Q}_{indep}$ is used to infer the rewards ($\hat{R}_{indep}$) of the independent agent. For the learning agent, a selfish value function ($Q_{selfish}$) is trained on rewards observed from the environment ($R$) and a sympathetic value function ($Q_{symp}$) is trained on sympathetic rewards ($R_{symp}$). $R_{symp}$ is a convex weighted sum of $R$ and $\hat{R}_{indep}$. The weighting is determined by the sympathy function $\beta(s,a)$ which outputs the degree of selfishness. $\beta(s,a)$ takes as inputs the value functions of the independent agent and selfish learning agent, and utilised a state prediction model $M(s,a)$. Actions are taken by the learning agent according to $Q_{symp}$}
  \label{fig:framework}
\end{figure}
\fi

We formulate our problem within the context of a Markov Decision Process (MDP) framework \cite{puterman2014markov} $\mathcal{<S,A,T,R,\gamma>}$ where $\mathcal{S}$ is the state space, $\mathcal{A}$ is the space of actions, $\mathcal{T} : \mathcal{S} \times \mathcal{A} \rightarrow \mathcal{S}$ is the transition function governing the probability of moving to the next state $s'$, having taken an action $a$ in the current state $s$. $\mathcal{R} : \mathcal{S} \times \mathcal{A} \rightarrow \mathbb{R}$ defines the reward an agent receives for taking an action in the current state, and $\gamma \in (0,1]$ is the discount factor.

To model the behaviour of the independent agent using EMOTE, we consider environment settings consisting of a learning agent, which we train, and one (or more) independent agent(s) which shares the same action and state spaces
%\thommen{actually I was thinking about it, and this method allows the transition dynamics of the two agents to be different, which is quite useful when you have a heterogeneous set of agents. You could include this as an advantage, but if there's no time/space right now, its fine - maybe we can add it later.} 
as the learning agent, and behaves as per a fixed policy. We assume that the underlying reward function of the independent agent is unknown to the learning agent. Our EMOTE architecture consists of training the action-value function for the learning agent $Q_{learn}$ using rewards $R$ returned from the environment, and simultaneously estimating the independent agent's action-value function as $\hat{Q}_{indep}$, using the latter's $<s_{i},a_{i},s_{i}^{'}>$ trajectories which we assume to be accessible (similar to a real world setting involving robots who share sensory input information). Each agent is assumed to also have their own reward function. 
%learning agent has access to the observations and actions of the independent agent, but the latter's underlying reward function is unknown to the former.
 %In order to estimate this reward function, the learning agent is assumed to have access to the independent agent's $<s_{i},a_{i},s_{i}^{'}>$ trajectories, derived from the independent agent's fixed policy. %Both agents are assumed to share the same action space. 
  %We propose the EMOTE architecture to model the independent agent's action-value function $\hat{Q}_{indep}$. 
  EMOTE uses relevant IRL methods to estimate the reward function $\hat{R}_{indep}$ of the independent agent from the estimated action-value function $\hat{Q}_{indep}$. Further details of the EMOTE architecture are described in the subsequent sections.

\begin{figure}[ht]
  \centering
  \includegraphics[width=0.9\linewidth]{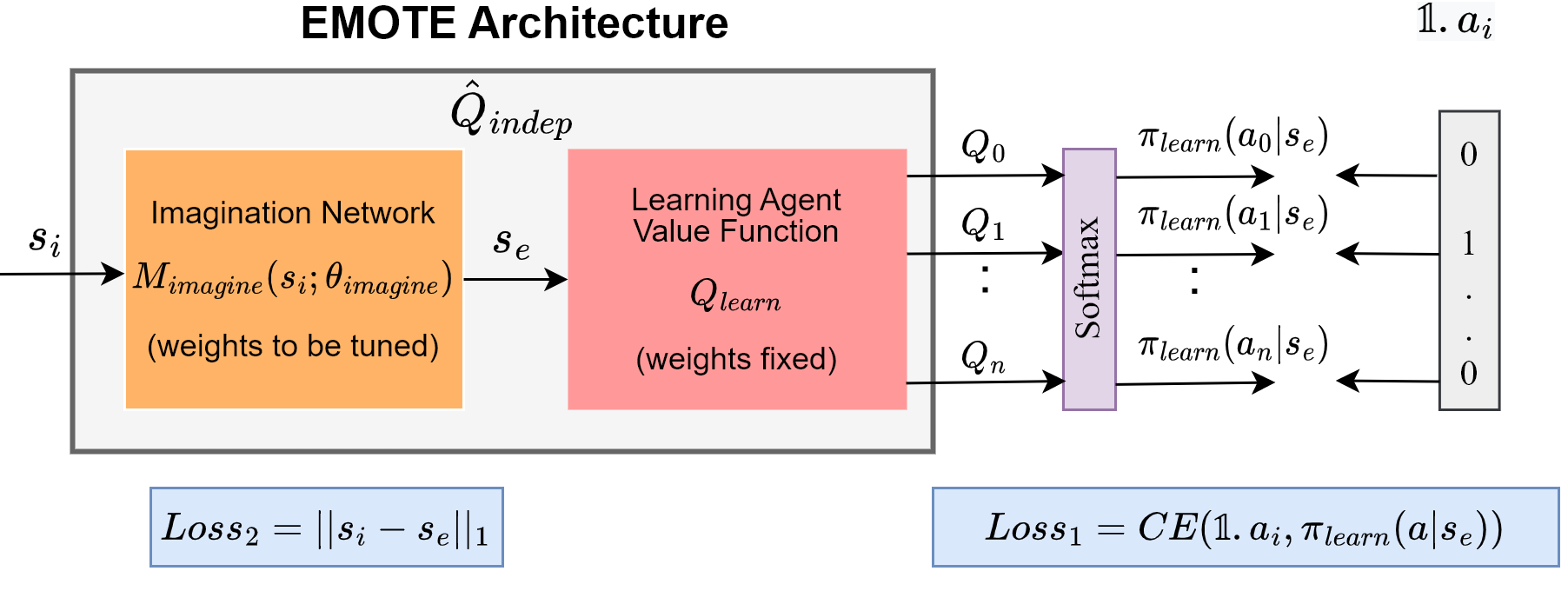}
  \caption{EMOTE architecture of the independent agent's value function $\hat{Q}_{indep}$ comprising a two stage neural network. The first, the Imagination Network $M_{imagine}$, takes in the state $s_{i}$ perceived by the independent agent and outputs an empathetic state $s_{e}$. The second is a copy of the learning agent's value function $Q_{learn}$. $s_{e}$ is fed into $Q_{learn}$ and associated Q-values are output. Only $M_{imagine}$ is trained via a loss function comprising the difference between $s_{i}$ and $s_{e}$ and the categorical cross entropy between the predicted softmax actions $\pi_{learn}(a|s)$ and the observed one hot encoded action of the independent agent $\mathbbm{1} \cdot a_{i}$.}  
  \label{fig:empathetic_valuefunction}
\vskip -0.2in
\end{figure}

\subsection{The EMOTE Architecture}\label{sec:emp-arch}
%\subsubsection{Empathy-based architecture}\label{sec:emp-arch}
%Sitting within the IRL module as seen in Figure \ref{fig:framework} the form of the independent agents empathy-based architecture value function is shown in more detail in Figure \ref{fig:empathetic_valuefunction}.
The components of the EMOTE architecture used to estimate the independent agent's value function $\hat{Q}_{indep}$ are shown in Figure \ref{fig:empathetic_valuefunction}. It consists of a two stage neural network architecture. The first of these, the Imagination Network ($M_{imagine}$) parameterised by $\theta_{imagine}$, takes state $s_{i}$, as observed by the independent agent, as input and outputs an empathetic state $s_{e}$ representing the independent agent's state as perceived empathetically by the learning agent. That is:
\begin{eqnarray}
s_{e} = M_{imagine}(s_{i};\theta_{imagine})
\label{eqn:IM_formula}
\end{eqnarray} 
Such that the learning agent's greedy action in $s_e$ matches the independent agent's action $a_i$. Formally, we define the empathetic state in Definition \ref{def:empstate}:
\begin{definition}[Empathetic State]
\label{def:empstate}
In a multiagent learning scenario involving a learning agent with action-value function $Q_{learn}$ and an independent agent (sharing same state space $\mathcal{S}$ and action space $\mathcal{A}$) who behaves per an arbitrary unknown policy, an empathetic state $s_e$ is a state where the learning agent's greedy action matches the independent agent's observed action $a_i$ in state $s_{i}$.

\begin{eqnarray*}
 \underset{a'}{argmax}{ 
 \; Q_{learn}(s_{e},a')} = a_{i}
\label{eqn:IM_condition}
\end{eqnarray*}
\end{definition}
%We aim to learn the transformation function $M_{imagine}$, where:
%if it were inputted into the learning agent's action-value function $Q_{learn}$, it would elicit a greedy action that matches the independent agent's action $a_{i}$ in state $s_{i}$. As such, we aim to learn the state transformation function $M_{imagine}$, where:

\begin{figure}[ht]
\vskip -0.2in
\begin{center}
\centerline{\includegraphics[width=0.5\columnwidth]{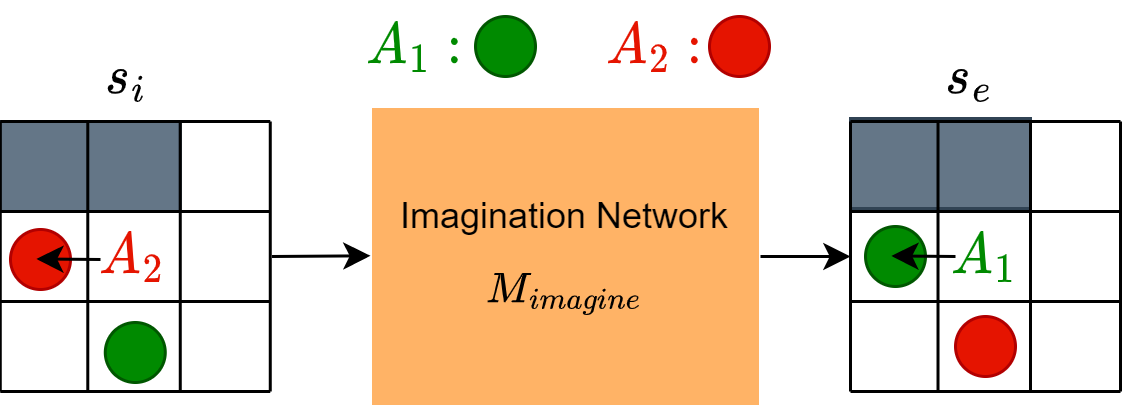}}
\caption{Example: Imagination Network. $s_{i}$: perceived by Agent 2. $s_e$: Agent 1's empathetic perception of $s_{i}$.}
\label{fig:emp_example}
\end{center}
\vskip -0.3in
\end{figure}

Our work is applicable when learning and independent agents share analogous features, and a sufficient degree of analogy exists. These are defined:
\begin{definition}[Analogous features]
 %\manisha{Are} specific features of the empathetic state that \manisha{when mapped and swapped with another feature,} induce the learning agent to mimic the independent agent's actions.
    A subset of features $f\subseteq \mathcal{F}$ contained in state $s\in\mathcal{S}$ is said to be analogous if there exists another subset of features $f_{e}\subseteq \mathcal{F}$, which would produce an empathetic state $s_e$ as defined in Definition \ref{def:empstate}, when $f$ is swapped with $f_e$, where $\mathcal{F}$ is the feature space.
\end{definition}
\begin{definition}[Degree of Analogy]
    The degree of analogy between two agents is the fraction of analogous features in the feature space.%\manisha{fraction of features which are analogous} \st{total number of analogous features} in the state space $\mathcal{S}$.
\end{definition}
%\thommen{Do analogous features by definition occur in pairs? Also, what about groups of features?}\manisha{not necessarily... there could be one to many mappings I guess. e.g. LP to IP or Floor}\thommen{Yes, but still pairs right? Eg: (LP,IP), (LP,Floor). I mean a feature cant be analogous without another feature. Might be worth writing it as analogous feature pair wherever appropriate. Also, I mentioned groups of features because it may be possible for a feature to not be analogous unless there are other features that are also changed along with that feature. Eg: key may not be analogous to access card by itself, but it will be analogous when the lock/access card reader features are also swapped. Maybe writing something about the limitation of this definition will be useful. Something like `Although Def 2 only considers single features, we acknowledge that feature groups can also be analogous...}
\paragraph{An Illustrative Example:} Figure \ref{fig:emp_example} shows an example where Agent 1 desires green pellets while Agent 2 desires red pellets. If Agent 2 observes state $s_{i}$, it would move left in order to obtain the red pellet. If Agent 1 were to observe state $s_{i}$ from Agent 2's position, it would not move left, but rather down, towards the green pellet. If however state $s_{i}$, is fed into a trained Imagination Network, the resulting empathetic state $s_{e}$ would see the position of the pellets (analogous features) be swapped, as depicted on the right side of Figure \ref{fig:emp_example}. As a result, when Agent 1 is presented the transformed state $s_{e}$ (instead of $s_{i}$), it would select the same action (moving left) as Agent 2 in state $s_{i}$. Hence, we can interpret that through $s_{e}$, Agent 1 understands that how it behaves towards and values green pellets is analogous to how Agent 2 behaves towards and values red pellets. 

%\thommen{My take on the intuition - can you check this? I feel like it is a better alternative to the first part of the next para. \emph{In order to estimate the independent agent's action values, we pass the obtained empathetic state $s_e$ through a copy of the learning agent's own action-value function $Q_{selfish}$. The intuition is that since $s_e$ for the learning agent is analogous to the independent agent's state $s_i$, it is reasonable to assume that for a given action, how the learning agent values $s_e$ is analogous to how the independent agent values $s_i$. Hence, passing $s_e$ through the learning agent's action-value function $Q_{selfish}$ produces analogous action-values for the independent agent, as seen from the learning agent's point of view. The resulting action-values then correspond to the learning agent's interpretation of the independent agent's action-values, based on its own action-values (and hence, its own experiences and environment interactions).}}

In order to estimate the independent agent's action values, we pass the obtained empathetic state $s_e$ through a copy of the learning agent's own action-value function $Q_{learn}$. The intuition is that since $s_e$ for the learning agent produces the same actions (behaviours) as the independent agent in state $s_i$, it is reasonable to assume that for a given action, how the independent agent values $s_i$ is analogous to how the learning agent values $s_e$. Hence, passing $s_e$ through the learning agent's action-value function $Q_{learn}$ produces action-values analogous to the independent agent's. The resulting action-values then correspond to the learning agent's interpretation of the independent agent's action-values, based on its own action-values (and hence, its own experiences and environment interactions).
%The second model in the architecture is simply a copy of the learning agent's action-value function $Q_{selfish}$. $s_{e}$ is fed into this model to obtain the corresponding Q-values for each possible action. These Q-values are then representative of the learning agent's estimate of the long term expected rewards from each action for the independent agent, generated by using its own reward and value function as a guide. %The intuition behind this is that
%, firstly, $Q_{selfish}$ is trained directly on rewards returned from the environment via the learning agent's trajectories.  By 
%The intuition behind this is that by querying the value associated with the empathetic state and the independent agent's observed action as per the learning agent's action value estimate $Q_{selfish}$, we elicit an associated Q-value response to the question, \textit{``What value does the learning agent estimate for the action taken by the independent agent if the former is empathising with the latter"}. 
In our previous example (Figure \ref{fig:emp_example}), via the empathetic state which swapped the colours of the pellets, the resulting Q-value for the action taken by Agent 2 (going left towards red) is similar to that of Agent 1 taking the same action in the empathetic state (going left towards the imagined green pellet). 

Having now obtained $\hat{Q}_{indep}$, we can infer the rewards of the independent agent $\hat{R}_{indep}$ through an existing IRL method, e.g. Cascaded Supervised Learning \cite{cascadedSuperIRL2013}. In this way $\hat{R}_{indep}$ will have a similar scale or magnitude as that of the learning agent's rewards $R$. This is particularly useful for MARL algorithms which make use of a composite value or reward function in their design.

\paragraph{Loss Terms}\label{sec:loss-term}
%\thommen{I think most of this first para can be removed.} \manisha{Not sure what should be removed...} The loss $\mathcal{L}$ used to train the EMOTE action value function comprises the convex combination of two loss terms (Equation \ref{eqn:Loss_Function}). It is important to note that Only the weights of the Imagination Network $M_{imagine}$ are tuned via the loss. As the second model corresponds to an identical copy of the learning agent's value function $Q_{learn}$,  these weights will be trained using DQN.

While the weights of $Q_{learn}$ are trained using DQN, we train $M_{imagine}$ using a loss term (Equation \ref{eqn:Loss_Function}) comprising two parts. The first ($Loss_{1}$) is a categorical cross entropy (CE) loss which minimises the difference between the softmax predicted action from the $Q_{learn}$ copy and the action ($a_{i}$) actually taken by the independent agent. %focuses on the accuracy of the softmax predicted action from the $Q_{learn}$ copy compared to the action actually taken by the independent agent ($a_{i}$). 
 %This is computed via categorical cross entropy (CE) is used for this term. 
This loss is designed to ensure the greedy action of $Q_{learn}$, for $s_{e}$, matches the independent agent's action (per Equation \ref{eqn:IM_formula}).

\begin{equation}
\label{eqn:Loss_Function}
\begin{aligned}\mathcal{L}(\theta_{imagine}) = (1-\delta)\underbrace{CE(\mathbbm{1}.a_{i},\pi_{learn}(a|s_{e}))}_{Loss_{1}}
+ \delta\underbrace{\|{s_{i}-s_{e}}\|_{1}}_{Loss_{2}}
\end{aligned}
\end{equation}
where
\begin{equation}
\pi_{learn}(a|s_{e})=\frac{e^{Q_{learn}(s_{e},a)}}{\sum_{a}e^{Q_{learn}(s_{e},a)}}
\end{equation}
and $s_{e}$ is obtained as per Equation \ref{eqn:IM_formula}.

The second loss term ($Loss_{2}$) focuses on state reconstruction, aiming to produce an empathetic state $s_{e}$ matching the original state $s_{i}$. Together, the goal is to produce an empathetic state $s_{e}$ through minimal changes to $s_{i}$, so that common features (e.g. walls or floors) are reconstructed, and differences between $s_{e}$ and $s_{i}$ reflect the analogous features  
%\st{reflective of those} 
needed to evoke empathy (driven by $Loss_{1}$). 
%The intention behind this is interpretability such that when $s^{e}$ is observed, it is representative of $s$. When used in combination with $Loss_{1}$ the objective of the loss term is to recreate $s^{e}$ to be as similar to $s$ (via $Loss_{2}$) as possible with any differences reflective of those changes which are needed to induce empathy (driven by $Loss_{1}$). In the example in Figure \ref{fig:emp_example} $Loss_{2}$ ensures common features such as floors and walls remain the same while $Loss_{1}$ drives the required change of swapping the pellet colour. 
Components used to construct the two loss terms are shown in Figure \ref{fig:empathetic_valuefunction}.

%The intuition is to induce the interpretability of the empathetic state for a user to peer into and gain insights into what the learning agent believes it has in common with the independent agent. 
The user specified hyperparameter $\delta \in [0,1]$ balances the importance of reconstructing the empathetic state $s_{e}$ to be as similar as possible to the original state $s_{i}$ (interpretability) and the accuracy of the learning agent's predicted actions in $s_e$. We note that in practice, a copy of $Q_{learn}$ for EMOTE is updated every few episodes (to maintain stability). %More discussion of this is included in Section \ref{sec:delta_hyper}

%\input{Paper_Sections/algorithm}

%\subsubsection{Adapting the Cascaded Supervised Learning method}\label{sec:Cascaded_IRL}
%\input{Paper_Sections/adapt_cascaded}

\section{Experiments}
%\iffalse
\begin{figure}[!h]
%\vskip -0.2in
     \centering
     \begin{subfigure}{0.18\textwidth}
         \centering
         \includegraphics[width=\textwidth]{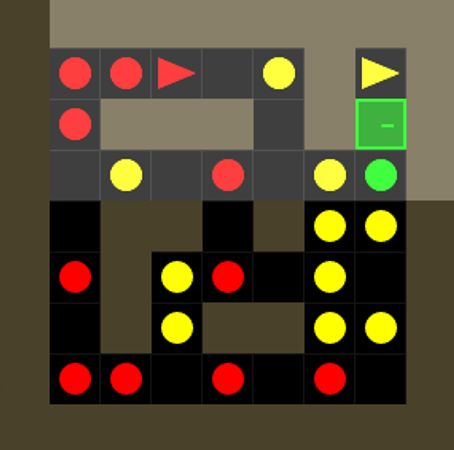}
         \caption{Ass. 1}
         \label{fig:y equals x}
     \end{subfigure}
     \begin{subfigure}[b]{0.18\textwidth}
         \centering
         \includegraphics[width=\textwidth]{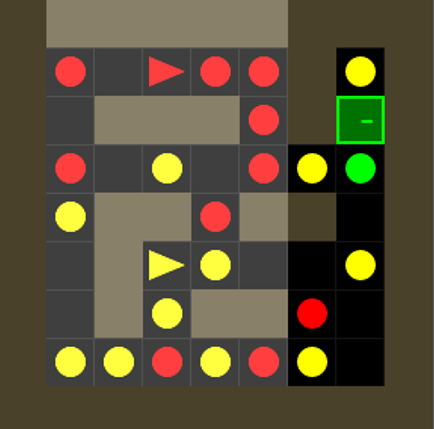}
         \caption{Ass. 2}
         \label{fig:three sin x}
     \end{subfigure}
     \begin{subfigure}[b]{0.18\textwidth}
         \centering
         \includegraphics[width=\textwidth]{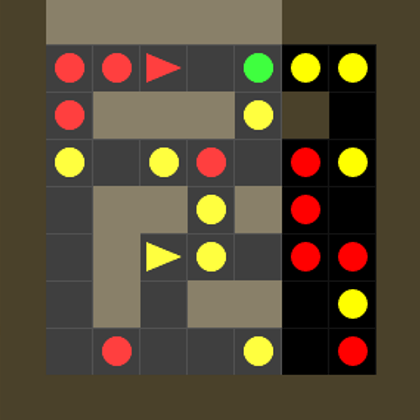}
         \caption{Adv. 1}
         \label{fig:three sin x}
     \end{subfigure}
     \begin{subfigure}[b]{0.18\textwidth}
         \centering
         \includegraphics[width=\textwidth]{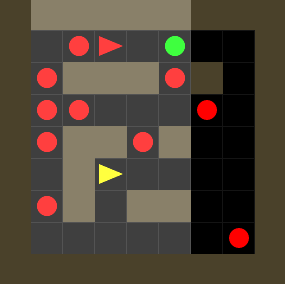}
         \caption{Adv. 2}
         \label{fig:three sin x}
     \end{subfigure}
        \caption{Learning agent (red arrow) and independent agent (yellow arrow) aim to collect their corresponding pellets. \textbf{Assistive env.} (a) - (b): Closed door (green square) can only be opened by learning agent pressing the green button (incurring a negative reward).  \textbf{Adversarial env.} (c) - (d): Independent agent can harm the learning agent. When the green button is pressed, learning agent can, for a finite time, harm independent agent, receiving a positive reward.}
        \label{fig:Environment_layout}
\vskip -0.2in
\end{figure}
%\fi

Our work is constrained to settings that meet the criteria outlined in Section 3.1. To the best of our knowledge, works that fit within these settings were those by Senadeera et al. \cite{SympathyPaper}, Bussmann et al. \cite{TowardsEmpathicDQN}, Noothigattu et al. \cite{Noothigattu2019} and Papoudakis et al. \cite{papoudakis2021agent}. 
%In this work, however, 
We demonstrate integration of our proposed framework on \cite{SympathyPaper} only,
%we only demonstrate our proposal on \cite{SympathyPaper}, 
as other works assumed the independent agent’s behaviour was (a) random \cite{TowardsEmpathicDQN}, (b) involved a complex switching policy to train the learning agent involving imitation learning via human example trajectories \cite{Noothigattu2019} or (c) did not have access to the independent agent’s trajectories during testing \cite{papoudakis2021agent}. The Sympathy Framework of \cite{SympathyPaper} consists of both a joint action-value function and reward function in its design, providing a useful illustration of the benefits proposed by the EMOTE architecture. To demonstrate the versatility of our proposed approach, we designed experiments that (1) illustrate various environment settings (two assistive and two adversarial), (2) demonstrate the potential of EMOTE to perform well even under differing degrees of the analogy assumption, (3) show the ability for the empathetic state to be constructed both as a feature-by-feature transformation, or as a whole state (image) transformation, and (4) demonstrate how in addition to visual features, our approach can handle non-visual features that influence the dynamics of the environment in a complex and non-linear fashion. %\thommen{Maybe we can emphasize this when explaining about the buttons? Do we need to include it here?}
%\st{Figure \ref{fig:Environment_layout} shows our four environment settings.} 
Environments were designed using MarlGrid \cite{MarlGrid} based on MiniGrid \cite{gym_minigrid} (code in Supplementary).

Played in finite, episodic environments where actions are taken sequentially by each agent, goal of learning agent (red arrow) is to complete its assigned task of collecting all red pellets. At times, the agent may be faced with situations where it has the ability to behave selfishly or even harm the independent agent (yellow arrow). The Sympathy Framework trains the learning agent to still complete its task whilst being considerate of the independent agent, even though there is no reward-driven incentive for doing so. This is done by training the learning agent on a \emph{sympathetic reward} \cite{SympathyPaper}, a convex weighted sum of the learning agent's reward, and the independent agent's inferred reward. The weighting is determined by a selfishness term, which is a function of both agents' action-value functions.

EMOTE replaces the network used to model the independent agent in \cite{SympathyPaper}, but we use the same IRL method (Cascaded Supervised Learning \cite{cascadedSuperIRL2013}) to infer rewards. This IRL method applies supervised learning to train the model using state-action pairs, following which an inversion of the Bellman Equation is applied to extract the reward function. EMOTE is assessed by three criteria:

%\thommen{Do we need all these details? We already mentioned they estimate the indep reward function with scaling etc., And we described the current approach above. So I think you can skip the details and just say we replace their reward estimation method with EMOTE. And then skip to `We assess EMOTE...'}Within the Framework is an IRL module made of two components: (1) a network to model the independent agent's action-value, and (2) a component to infer the independent agent's reward based on this action-value function. The IRL module follows the Cascaded Supervised Learning method proposed by \cite{cascadedSuperIRL2013}. 

\begin{enumerate}
    \item \textbf{Performance:} Whether with EMOTE, the learning agent behaves considerately towards the independent agent while completing its own task.
    \item \textbf{Inferred Independent Agent Rewards:} Are the rewards inferred via IRL from the EMOTE architecture comparable to the learning agent's rewards?%, and how does it compare to those inferred by \cite{SympathyPaper}? %\thommen{do we need `..and how does it compare to those inferred by \cite{SympathyPaper}?'? I feel like it'll seem like we are specifically targeting this paper}\manisha{shall we just cut out the bit from '..and how does it compare...' onwards?}
    \item \textbf{Explainability}: Are the analogous features and associated empathetic states informative and reflective of the independent agent's behaviour?
    %and does it correspond to the expected nature (eg: highly positive, negative, neutral etc.,) of empathetic rewards?\thommen{the part about empathetic state is okay, but isn't the rewards part more related to point 2. (although I see why you put it under explainability)?}
\end{enumerate}

\subsection{The games}

\paragraph{Assistive 1 and 2}
Both agents try to collect their corresponding coloured pellets. When the learning agent consumes one of its pellets, it receives +10 points. In Assistive 1 the independent agent is locked behind a door and in Assistive 2 one of the independent agent's pellets is locked behind the door. The door can only be opened by the learning agent stepping on the green button, which inflicts a small negative reward (-1) on the learning agent. Once the button is pressed, the door remains open for the remainder of the episode. The learning agent also receives an additional bonus reward (+5) when it `wins' the game by consuming all of its pellets.
%does not need to assist the independent agent in order to win the game (+5). 
In the absence of other rewards a step penalty (-1) applies for each step taken. Pellets are placed randomly in each episode and the episode ends if all learning-agent pellets are consumed or when the game timer runs out.

Ideally, a considerate learning agent would open the door and assist the independent agent, despite not being necessary for winning. These games are considered to have a moderate degree of analogy, as both agents react similarly to pellets, but differ as only the learning agent can open the door.
%, and even incurring a small negative reward for doing so. %We note that in all experiments, the independent agent's behaviour is simulated using a pretrained, fixed $\epsilon$-greedy policy.

\paragraph{Adversarial 1 and 2}
In Adversarial 1 both agents again try to collect their respective pellets, and the learning agent earns +20 points per pellet. In Adversarial 2 only the learning agent is concerned with pellet collection.  When the game starts the independent agent can harm the learning agent (resulting in -50 points for the latter and the game ending). If the learning agent steps on the green button however (switching the button status from 0 to 1), for a finite period of time, it can harm the independent agent, resulting in a positive reward (+10) for the learning agent. Harming occurs when the harming agent is within 1 square from the other agent. We investigate whether the learning agent, by virtue of the empathetic architecture, avoids harming the independent agent despite receiving positive environment rewards for doing so. Each episode terminates when the learning agent collects all pellets (receiving +30 reward), if the game timer runs out, or the learning agent is harmed. Adversarial 1 has a high degree of analogy as both agents have comparable reactions to elements (pellets and harming), whilst Adversarial 2 has a lower degree, as the only analogous feature is the ability to harm each other.

%In this environment, as both the vision field and status of the button are relevant to decision-making, we performed two variants of experiments- 

%The first had the button status manually switched (Emp(Feature)-B and Emp(Image)-B) (and the vision field trained). In this run, we assume that how the learning agent feels when the button status is 0, is how the independent agent feels when the button status is 1 (and vice-versa). In the second set of runs we trained both the vision field and a model to predict an empathetic button status value (Emp(Feature) and Emp(Image)). A time penalty of -1 for each step is applied.

\subsection{Baselines}
In each experiment, policies are learned via DQN \cite{mnih2015human}. A 5x5 field of vision is imposed around each agent, representing its visual state information. We compare EMOTE with the following baselines:

\emph{Selfish:} A selfish version of the learning agent which only has access to its own rewards from the environment (no modelling of the independent agent).

\emph{Sympathy:} Agents trained as per \cite{SympathyPaper}.

\emph{E-Feature:} A feature-based Imagination Model where each state cell is represented as a feature. $s_{e}$ is constructed with a feature-by-feature transformation.

\emph{E-Image:} An Image-based Imagination Model, where the entire state is transformed to create $s_{e}$.

\emph{Benchmark 1 - Swap (B-Vis):} The Imagination Model is replaced with a rule-based state transformation, such that the colours of the two agents' pellets are swapped. This mimics an oracle baseline that presumes the hypothesis - how the learning agent feels about its pellets is how the independent agent feels towards its own pellets. 

\emph{Benchmark 2 - Swap (B-Invis):} The Imagination Model is replaced with a rule-based state transformation which swaps the colours of the agent's pellets (like Benchmark 1) and additionally, makes the button invisible. The invisibility applies only to the observed state, and the button status remains. This mimics an oracle corresponding to the hypothesis that in addition to Benchmark 1's belief, the independent agent does not consider the button to be important (treats it the same as the floor), as it cannot press it.

%\textbf{Benchmark 3 - Swap (B-Recolor)}: A rule-based state transformation which swaps the colors of the agent's pellets. Also, the button color is changed to the learning agent pellet's color. This mimics the hypothesis that if the button was the same color as the learning agent's pellet, it would incentivise the door to be opened.

\subsection{Performance}
\begin{table*}
\vskip -0.3in
\begin{center}
\caption{Performance results. Assistive games show win rate (higher is better) and door-opening rate (higher is better). Adversarial games show win rate (higher is better) and harm rate (lower is better) from the learning agent toward the independent agent. Cells are shaded green where performance exceeded Sympathy. Cells have bold text where intended considerate behaviour or comparable win rates were achieved, compared to Selfish run.}
\vskip 0.15in
\begin{tabular}{cccccccc}
\toprule
& & \multicolumn{1}{c}{Selfish} & \multicolumn{1}{c}{Sympathy} & \multicolumn{1}{c}{E-Feature} & \multicolumn{1}{c}{E-Image} & \multicolumn{1}{c}{B-Vis} & \multicolumn{1}{c}{B-Invis} \\ \midrule 
& Door & 0.44$\pm$0.15 & \textbf{0.99$\pm$0.01} & \cellcolor[HTML]{9AC68B}\textbf{0.99$\pm$0.01} & \textbf{0.83$\pm$0.11} & \cellcolor[HTML]{9AC68B}\textbf{1$\pm$0} & \cellcolor[HTML]{9AC68B}\textbf{1$\pm$0} \\ 
Ass. 1 & Win & 0.8$\pm$0.11 & \textbf{0.93$\pm$0.05} & \cellcolor[HTML]{9AC68B}\textbf{0.93$\pm$0.05} & \textbf{0.92$\pm$0.06} & \textbf{0.92$\pm$0.06} & \cellcolor[HTML]{9AC68B}\textbf{0.93$\pm$0.05} \\ \midrule 
 & Door & 0.09$\pm$0.07 & \textbf{0.17$\pm$0.1} & \cellcolor[HTML]{9AC68B}\textbf{0.4$\pm$0.15} & \cellcolor[HTML]{9AC68B}\textbf{0.38$\pm$0.15} & \cellcolor[HTML]{9AC68B}\textbf{0.21$\pm$0.11} & \cellcolor[HTML]{9AC68B}\textbf{0.19$\pm$0.11} \\ 
Ass. 2 & Win & 0.89$\pm$0.08 & \textbf{0.92$\pm$0.06} & \cellcolor[HTML]{9AC68B}\textbf{0.94$\pm$0.05} & \cellcolor[HTML]{9AC68B}\textbf{0.92$\pm$0.06} & \textbf{0.89$\pm$0.08} & \cellcolor[HTML]{9AC68B}\textbf{0.92$\pm$0.06} \\ \midrule  
 & Harm & 0.5$\pm$0.15 & \textbf{0.45$\pm$0.15} & \cellcolor[HTML]{9AC68B}\textbf{0.24$\pm$0.12} & \cellcolor[HTML]{9AC68B}\textbf{0.17$\pm$0.1} & \cellcolor[HTML]{9AC68B}\textbf{0.1$\pm$0.07} & \cellcolor[HTML]{9AC68B}\textbf{0.1$\pm$0.07} \\  
Adv. 1 & Win & 0.46$\pm$0.15 & 0.26$\pm$0.13 & \cellcolor[HTML]{9AC68B}0.4$\pm$0.15 & \cellcolor[HTML]{9AC68B}0.41$\pm$0.15 & \cellcolor[HTML]{9AC68B}0.37$\pm$0.15 & \cellcolor[HTML]{9AC68B}0.34$\pm$0.14 \\ \midrule
 & Harm & 0.29$\pm$0.13 & \textbf{0.12$\pm$0.09} & \cellcolor[HTML]{9AC68B}\textbf{0.1$\pm$0.08} & \cellcolor[HTML]{9AC68B}\textbf{0.12$\pm$0.09} & \cellcolor[HTML]{9AC68B}\textbf{0.05$\pm$0.04} & \cellcolor[HTML]{9AC68B}\textbf{0.03$\pm$0.02} \\ 
Adv. 2 & Win & 0.73$\pm$0.13 & 0.31$\pm$0.14 & \cellcolor[HTML]{9AC68B}0.44$\pm$0.15 & \cellcolor[HTML]{9AC68B}0.47$\pm$0.15 & \cellcolor[HTML]{9AC68B}0.34$\pm$0.14 & \cellcolor[HTML]{9AC68B}0.33$\pm$0.14 \\\midrule
\end{tabular}
\label{table:win_harm_door}
\end{center}
\vskip -0.3in
\end{table*}

Table \ref{table:win_harm_door} shows the Win Rate and Door Open rate for the two Assistive environments, as well as the Win Rate and Harm Rate (rate at which the learning agent harms the independent agent) for each of the adversarial experiments. Cells highlighted in green show where the EMOTE architecture or \emph{Benchmarks} had similar or better performance than the \emph{Sympathy} baseline. Cells with bold text indicate better or similar performance to the \emph{Selfish} baseline. 

In the Assistive environments, the \emph{E-Feature} and \emph{E-Image} EMOTE baselines result in high win and door open rates, producing similar or better results to the \emph{Sympathy} baseline, with it at times even outperforming the \emph{Benchmarks}. In adversarial environments, all harm rates were lower than the \emph{Selfish} baseline. \emph{E-Feature}, \emph{E-Image} and the \emph{Benchmarks} outperformed \emph{Sympathy} (state-of-the-art baseline) with a lower harm rate, and produced higher win rates. The \emph{Benchmarks} produced lower harm rates than EMOTE however. The win rates of other baselines compared to \emph{Selfish} were either similar (Adversarial 1) or lower (Adversarial 2).  Overall, EMOTE induced more considerate behaviours than \emph{Sympathy}.
%(harming agent As a result of this, the learning agent infers that how the independent agent behaves when the button status is active (when the learning agent has the ability to harm) is much the same as how itself feels when the independent agent can harm. The reverse is true for situations when the button status is inactive.

\label{sec:performance}

\subsection{Inferred Reward Values}
\begin{figure*}[ht]
\vskip -0.2in
\begin{center}
\centerline{\begin{subfigure}[b]{0.22\textwidth}
         \centering
         \includegraphics[width=\textwidth]{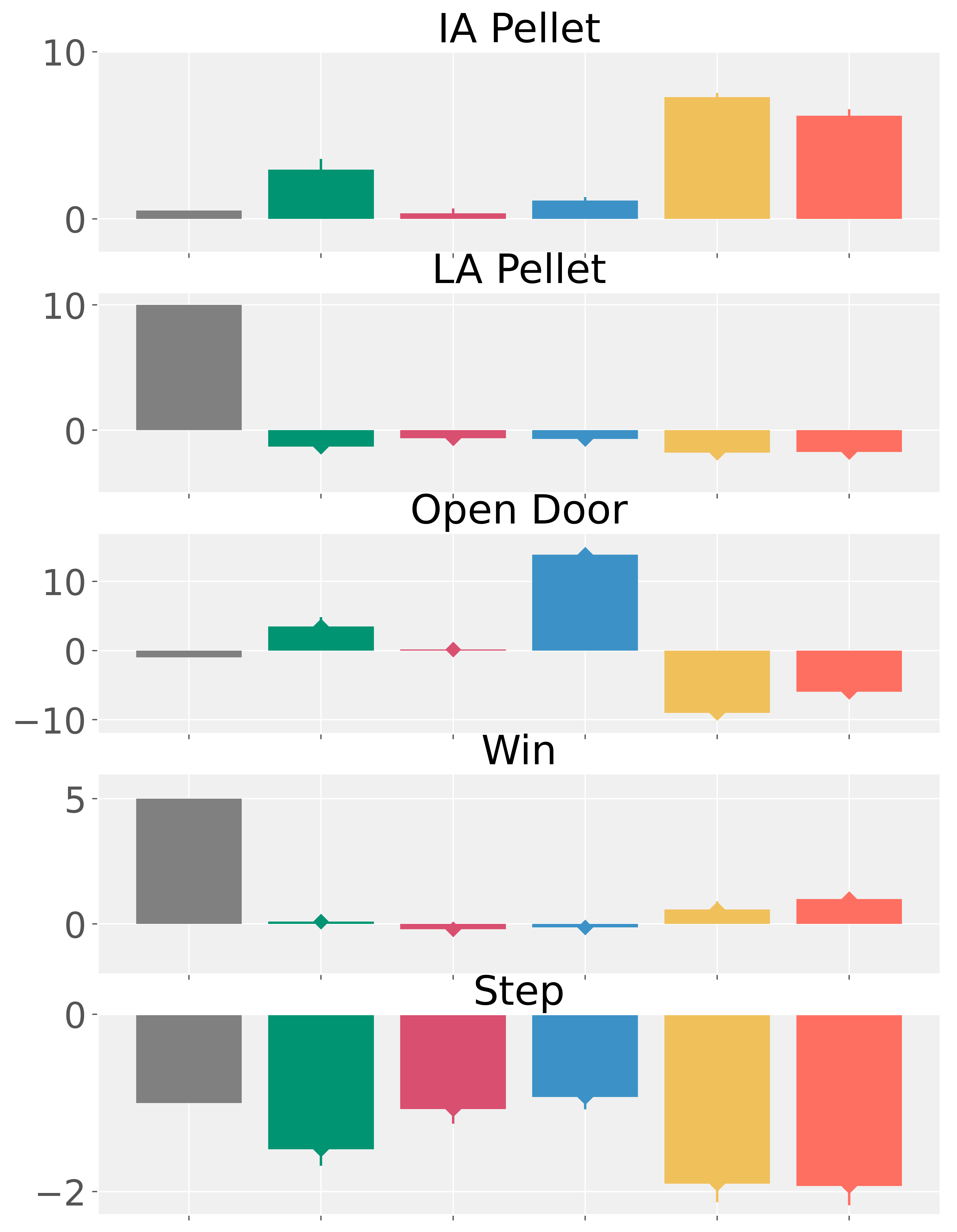}
         \caption{Assistive 1}
     \end{subfigure}
     \begin{subfigure}[b]{0.22\textwidth}
         \centering
         \includegraphics[width=\textwidth]{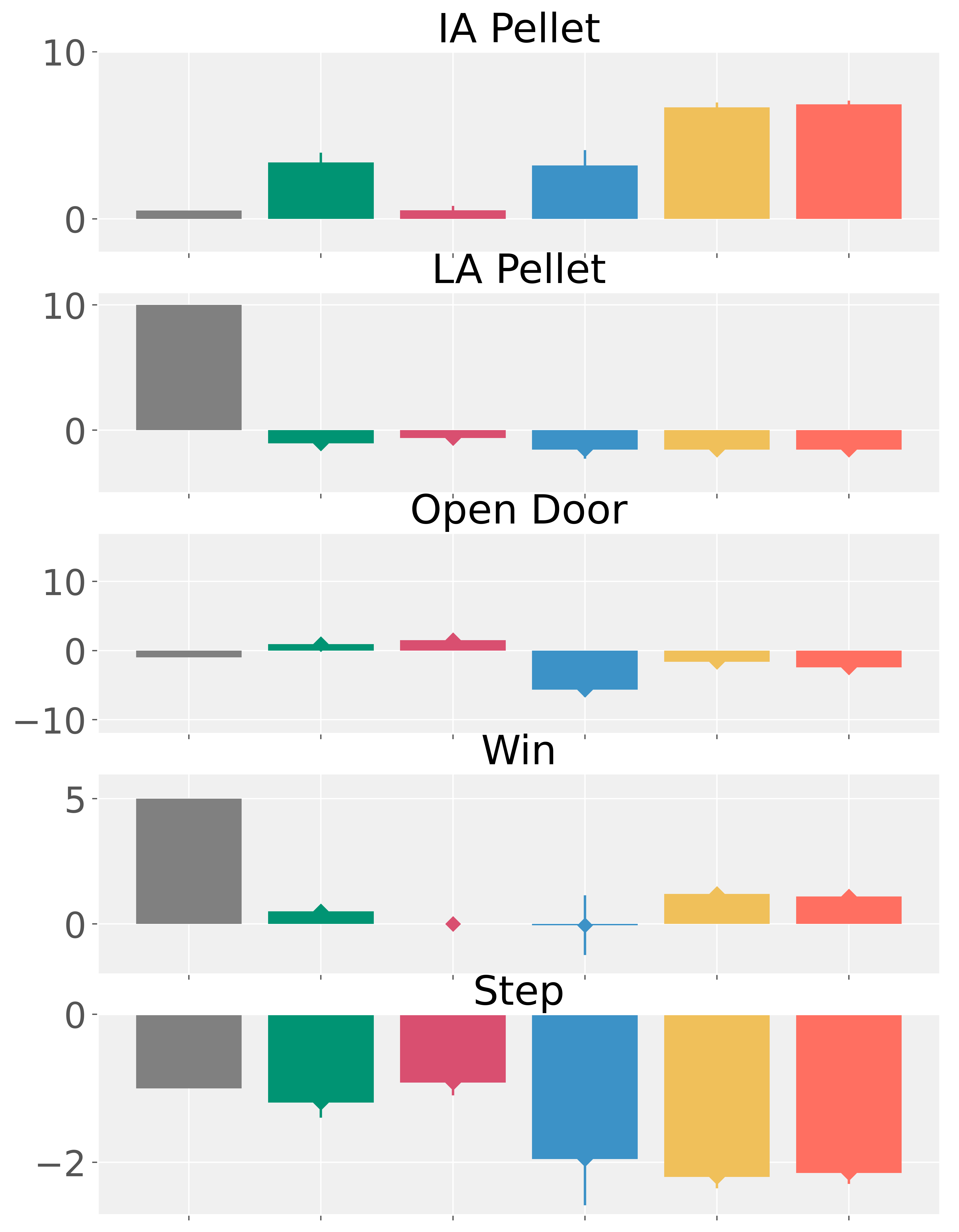}
         \caption{Assistive 2}
     \end{subfigure}
     \begin{subfigure}[b]{0.215\textwidth}
         \centering
         \includegraphics[width=\textwidth]{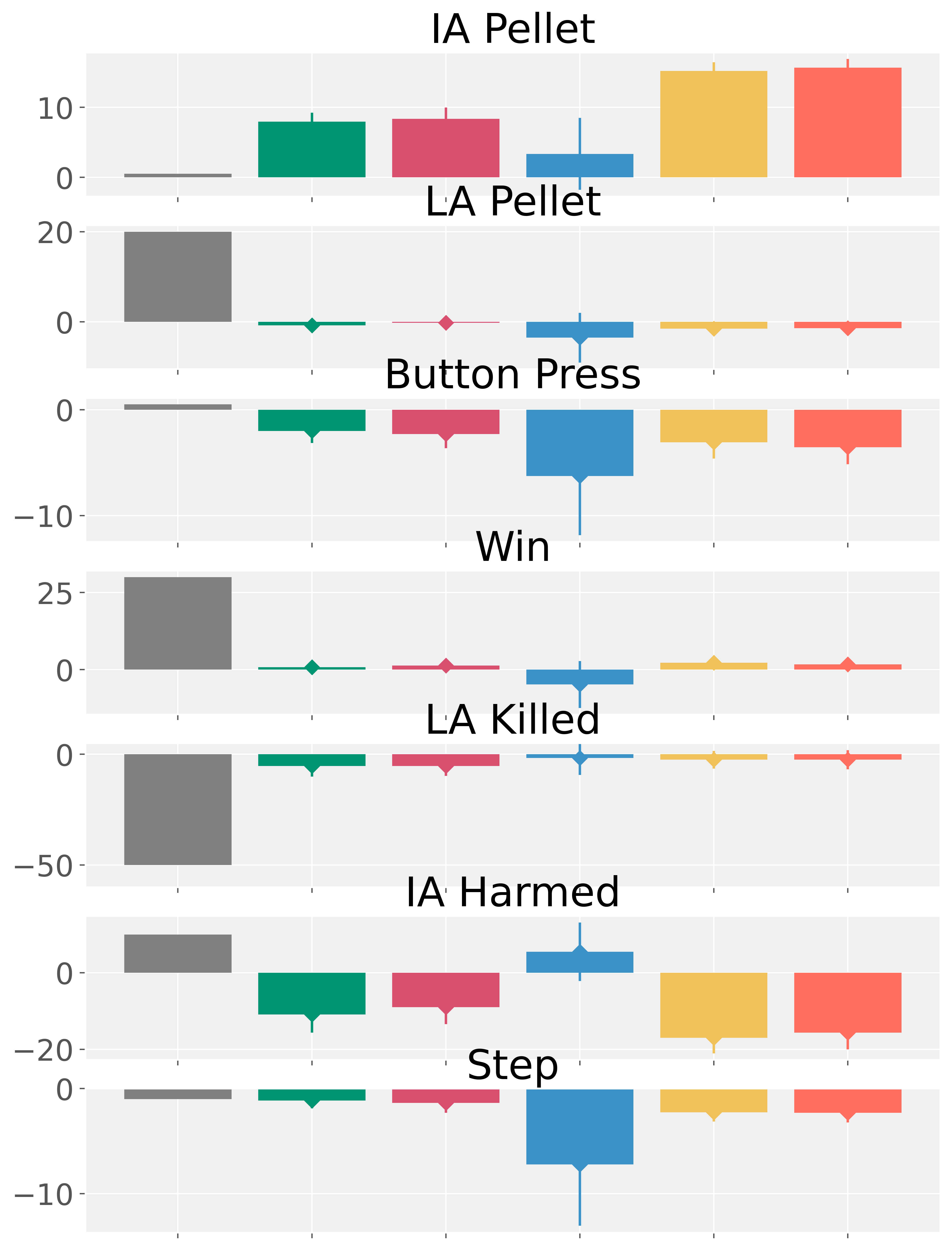}
         \caption{Adversarial 1}
     \end{subfigure}
     \begin{subfigure}[b]{0.31\textwidth}
         \centering
         \includegraphics[width=\textwidth]{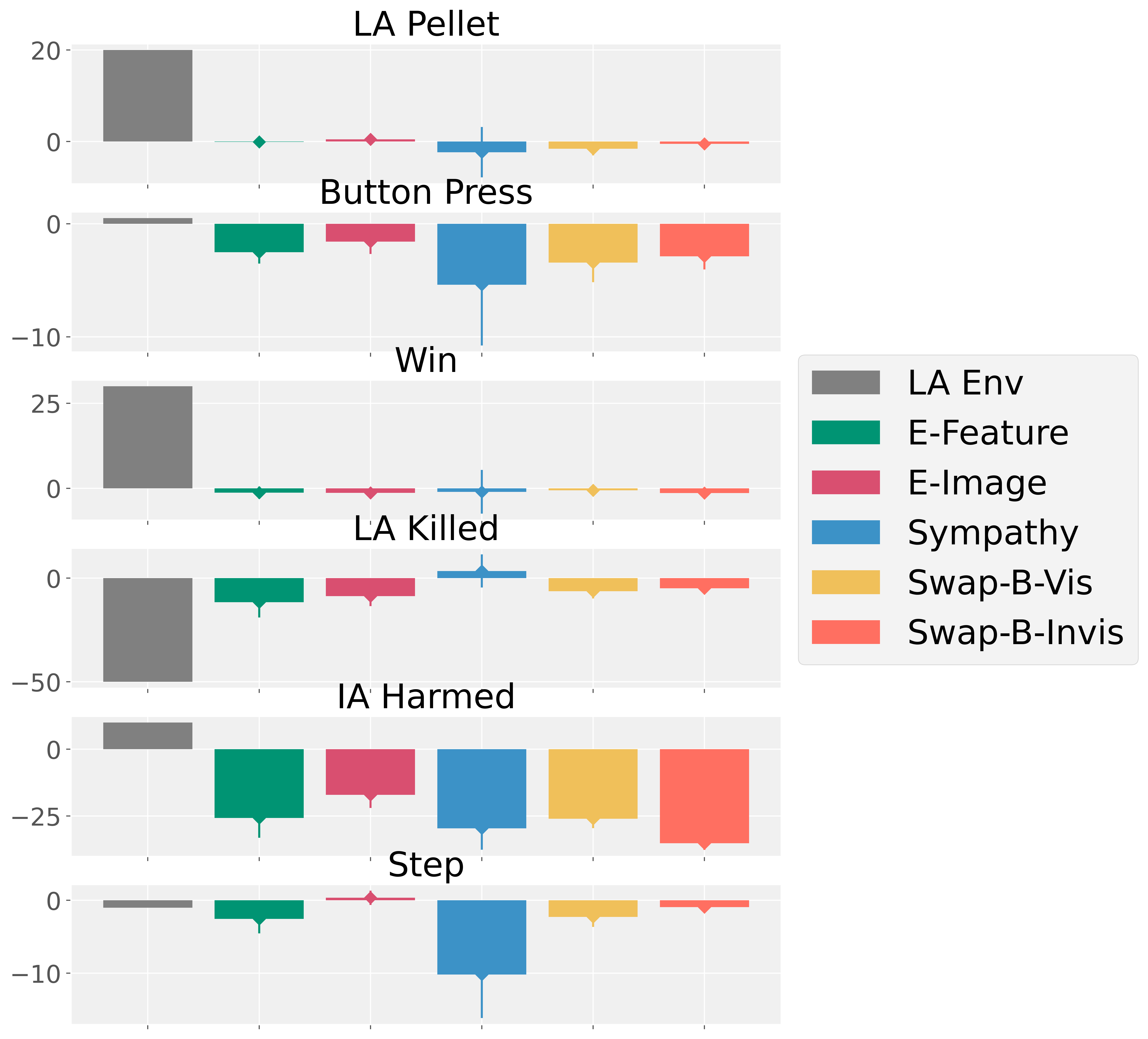}
         \caption{Adversarial 2}
     \end{subfigure}}
\caption{Independent Agent's (IA) $\hat{R}_{indep}$ (averaged over last 100 episodes) derived from $\hat{Q}_{indep}$. Learning Agent's (LA) environment rewards are included for reference. \emph{Sympathy} rewards are scaled to have the same $l1$ norm as the LA's rewards.}
\label{fig:IRL_results}
\end{center}
\vskip -0.3in
\end{figure*}

Figure \ref{fig:IRL_results} shows the independent agent's inferred rewards ($\hat{R}_{indep}$) from the EMOTE and \emph{Benchmark} action-value functions ($\hat{Q}_{indep}$), and contrasts them against those from \emph{Sympathy}. For reference the learning agent's environmental rewards for the same features are shown alongside the independent agent's $\hat{R}_{indep}$.
% We examine the independent agent's inferred rewards ($\hat{R}_{indep}$) from EMOTE and \emph{Benchmarks}, and contrast against those from \emph{Sympathy}, shown in Figure \ref{fig:IRL_results} (averaged over last 100 episodes). For reference the learning agent's environmental rewards for the same reward features are shown alongside the inferred rewards for the independent agent.
% , where the learning agent's reward for ``LA Pellet'' should, for instance, be comparable to the independent agent's reward for ``IA Pellet''.
% To assist with assessing comparability of the inferred rewards, the learning agent's environmental rewards are also included.

\subsubsection{Assistive}
To judge whether the inferred independent-agent's rewards are comparable to the learning agent's environmental rewards we can examine similarities in reward values across various features. 
%\emph{we need to look from each agent's perspective.} 
For instance one would expect the baselines' inference for the independent agent consuming its pellet (IA pellet) to look similar to the learning agent's environmental reward for consuming its own pellet (LA pellet). 
In Figure \ref{fig:IRL_results} (a) and (b) (Assistive 1 and 2) \emph{E-Feature}, \emph{E-Image}, and the \emph{Benchmark} baselines all infer positive rewards for IA pellet, though magnitudes vary. The rewards inferred by the \emph{Benchmarks} are closest to the learning agent's reward for LA pellet (+10).
%that the learning agent receives for consuming its own pellet (LA pellet). 
Opening the door elicited a slight positive value in the EMOTE runs, and the value for taking a step was close to that of the learning agent's step value (-1). Under the EMOTE and \emph{Benchmark} baselines, the rewards inferred for Assistive 1 and 2 are similar. %This highlights the potential of the architecture to learn a robust model demonstrated by consistent reward values even when the layout of the environment changes. 
This demonstrates the potential of the architecture to learn a robust model which infers consistent reward values even when the layout of the environment changes. 
In contrast, \emph{Sympathy} inferred different rewards for the two environments. It inferred a strong positive value for the button being pressed in Assistive 1, but a slight negative value in Assistive 2 (Figure \ref{fig:IRL_results} a) and b)). Additionally despite the high win rate in Assistive 2, the \emph{Sympathy} baseline did not result in as high of a door opening rate (Table \ref{table:win_harm_door}) as it wrongly inferred a negative reward for door opening. %\st{This is due to the unexpected negative value inferred for the door being opened.}\thommen{Isnt it understood that it is because of the negative value? To me, the takeaway, which should replace this sentence should be that the empathy baselines are able to exhibit more considerate behaviours.}\manisha{The considerate behaviours comment is more applicable in the 4.3 performance section? This section is just IRL} 
 All baselines inferred the independent agent associates a value close to 0 for the learning agent consuming a pellet.

\subsubsection{Adversarial}

For the adversarial games (Figure \ref{fig:IRL_results} (c) and (d)) \emph{E-Feature}, \emph{E-Image} and both \emph{Benchmarks} were able to capture the strong negative reward that the independent agent associates with being harmed. This is similar to the environmental reward the learning agent receives when it is killed. In Adversarial 1, a positive value was also associated with the independent agent consuming its own pellet. In Adversarial 1, the \emph{Sympathy} baseline was not able to infer a negative reward for the independent agent  being harmed. It also failed to capture a strong enough positive reward for the independent agent consuming a pellet and inferred a strong negative reward for taking a step and pressing the button, leading to a low win rate and high harm rate, similar to that of \emph{Selfish} (Table \ref{table:win_harm_door}). 
%As a result of this incorrect reward inference the win rate was adversely affected (Table \ref{table:win_harm_door})
We expect the poor performance is due to the order of magnitude difference, resulting from the $l1$ norm scaling, between rewards inferred for taking a step and button press relative to the other reward features. This inappropriate scaling of the reward components in turn results in poor estimates of the sympathetic reward.
%\manisha{When combined to construct the sympathetic rewards, the learning agent is biased to a false view of the independent agent.}
%, illustrating the fragility of using $l1$ norm scaling
%\thommen{As a result, the $l1$ norm scaling is affected, leading to an innacurate model of the independent agent.} %
 In Adversarial 2, the \emph{Sympathy} baseline inferred a strong negative for independent agent being harmed, leading to a reduced harm rate. Despite this \emph{Sympathy} had lower win rates compared to EMOTE or \emph{Benchmarks} (Table \ref{table:win_harm_door}).
%once again inferred a strong negative reward for taking a step, and despite the negative values it inferred for the independent agent being harmed, it lead to lower win rates compared to EMOTE or the \emph{Benchmarks} (Table \ref{table:win_harm_door}). 
EMOTE produced consistent reward inferences, almost matching \emph{Benchmark}, unlike \emph{Sympathy} where rewards inferred varied by the environment configuration.
%\st{Once again EMOTE and benchmark runs}\thommen{should we highlight the benchmark runs? Isnt it more like an oracle baseline? Maybe you could say that EMOTE was able to produce consistent reward inferences, almost matching the ones obtained through the benchmark (oracle) runs, unlike the Sympathy baseline in which the rewards inferred varied considerably depending on the environment configuration.}\manisha{ok. added} \st{were able to produce consistent reward inferences between the adversarial games, whilst Sympathy was not.}

\subsection{Empathetic State}
\begin{figure*}[ht]
\vskip -0.2in
\begin{center}
\centerline{\includegraphics[width=0.95\textwidth]{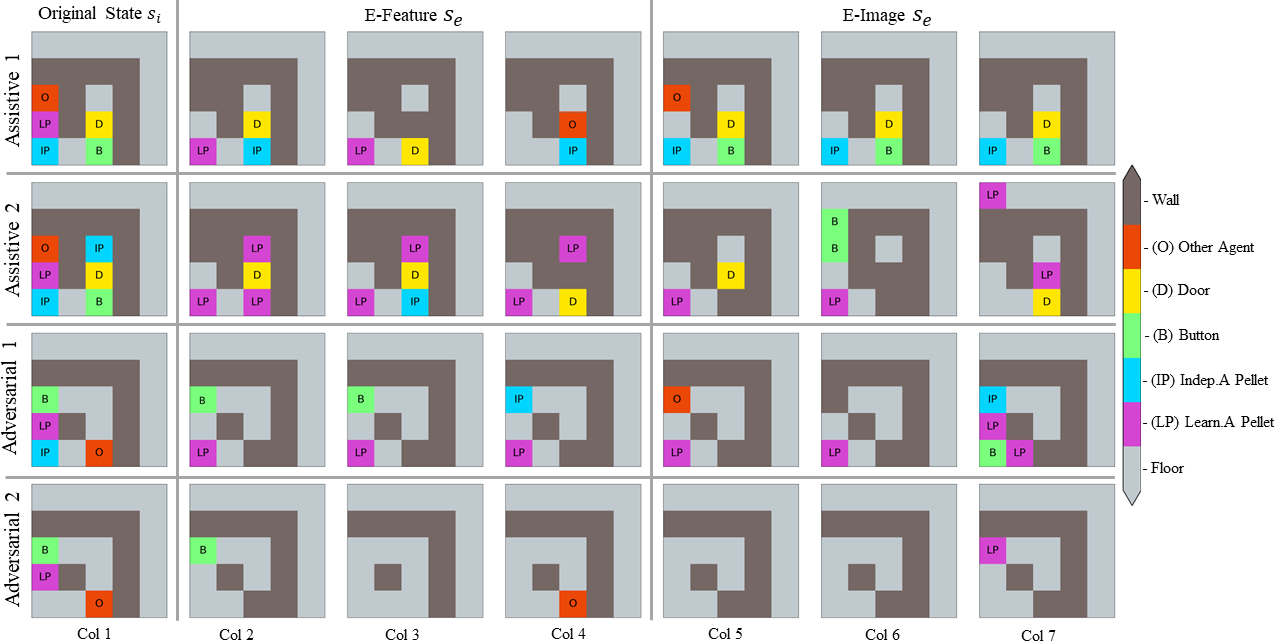}}
\caption{Empathetic states $s_{e}$ produced by E-Feature (Col 2 - 4) and E-Image (Col 5 - 7) for original state $s_{i}$ (Col 1). Details on construction of $s_{e}$ plots are provided in the Supplementary.}
\label{fig:imagined_state}
\end{center}
\vskip -0.4in
\end{figure*}

EMOTE's key benefit is its ability to produce a human-interpretable empathetic state $s_{e}$ which can be used to explain some of the performance results from Section \ref{sec:performance}. Figure \ref{fig:imagined_state} shows an original state $s_{i}$ for each of the four environments examined alongside three examples of the final empathetic state $s_{e}$ (at the end of training) generated from the \emph{E-Feature} and \emph{E-Image} Imagination Networks.
%For the 4 environments examined, we output $s_{e}$ corresponding to the original states $s_{i}$. These are shown in Figure \ref{fig:imagined_state} with $s_{i}$ for each game alongside 3 examples of the final empathetic-state $s_{e}$ (at the end of training) generated from the \emph{E-Feature} and \emph{E-Image} Imagination networks. 
More final empathetic states, as well as examples of the change in the empathetic state for those shown in Figure \ref{fig:imagined_state} during training can be found in the Supplementary.

The learning agent's pellet (LP) is fairly consistently transformed to the floor by $s_{e}$, indicating the lack of importance the independent agent places on it.  We observed EMOTE outperforming \emph{Benchmarks} (Section \ref{sec:performance}). We infer that because the learning agent will have experienced more instances of the floor (relative to IP, which also disappear when consumed), this feature is better estimated in its value function, leading to better performance by reimagining the LP as floor. 
%\thommen{On the other hand, in the \emph{Benchmarks}, the action-value estimates around the feature of IP may be slow to converge, owing to the learning agent's limited interactions with this feature, especially during the initial stages of learning.}   
The colour of the independent agent's pellet (IP) becomes that of the learning agent's pellets (blue) in the transformed state $s_e$ thus explaining the inferred rewards for IA Pellet in Figure \ref{fig:IRL_results}. This suggests that the empathy architecture allows the learning agent to interpret the relationship between independent agent and its pellets as analogous to that between the learning agent and its own pellets.%much the same as what the independent agent feels towards its own pellet. 

In the empathetic states for Assistive 1 in Figure \ref{fig:imagined_state}, the button either remains unchanged, changes to a door or an independent agent's pellet. The independent agent does not value the button as it cannot interact with it thus it is mapped to features that are similarly irrelevant in the learning agent's own reward function.
%This is understandable as the learning agent is not attracted toward these features according to its own reward function. This is similar to how the independent agent behaves towards the button (due to its inability to get close to the button because it is stuck behind the door). 
In contrast, for the Assistive 2 environment, Figure \ref{fig:imagined_state} shows the button transforming at times to the learning agent's pellet. This is expected as how the independent agent moves towards the door (in front of which the button is placed) is similar to how the learning agent moves towards its own pellet. 

In Adversarial environments, the button usually disappears in the empathetic states. This is expected, as the independent agent has no influence over the button, and is thus not a feature of importance. However, it is important to observe the value of the predicted button status in the empathetic states, the results of which are shown in Table \ref{table:button_status}. For a button status value of 0 in $s_{i}$ (button is inactive and independent agent can harm the learning agent) the resulting status prediction in $s_{e}$ is closer to 1, while a button status value of 1 in $s_{i}$ (button is active and learning agent can harm the independent agent) produces a prediction in $s_{e}$ closer to 0. This indicates that the EMOTE architecture is able to associate the button status (a non-visual feature) with the power dynamics between agents, namely how the independent agent behaves when the button status is 1 (i.e. trying to avoid being harmed) is similar to how the learning agent behaves when the button status is 0 (i.e. also trying to avoid being harmed). In the adversarial environments, the \emph{Benchmarks} outperformed EMOTE for the harm metric as shown in Table \ref{table:button_status}. This is because the \emph{Benchmarks} are oracle baselines, where the button status was swapped using manually encoded rules. However, for EMOTE this was a difficult transformation (with a non-linear influence on the environment dynamics), that had to be learned (Table \ref{table:button_status}).
%\thommen{Here maybe you can emphasize the point about EMOTE being more than just visual features}

%(is able to infer that that is, when the learning agent is in a position to harm, the inference of the independent agent's association is that of the learning agent when it is in a vulnerable position)\thommen{this part in the brackets is confusing, and the sentence structure is strange. What did you want to say?}. 

\begin{table}
\vskip -0.3in
\begin{center}
\caption{Adversarial: Predicted $s_e$ button status for $s_i$ button status values
%$s_{e}$ button status in Adversarial games \thommen{is perceived inversely as the true button status ($s_e$ is close to $1$ when $s_i=0$ and vice-versa), indicating the correct inference of agent power dynamics.} %\manisha{$s_{i}$ button value of 0 (inactive) indicates independent agent can harm learning agent. Value of 1 (active) indicates learning agent can harm independent agent.}
}
%\vskip 0.15in
\begin{tabular}{lcccc}\toprule
&  \multicolumn{2}{c}{$s_{i}$ E-Feature} & \multicolumn{2}{c}{$s_{i}$ E-Image}
\\\cmidrule(lr){2-3} \cmidrule(lr){4-5}
 Game  & 0 & 1 & 0 & 1 \\\midrule
Adv 1      & 0.76 $\pm$ 0.11   & 0.24 $\pm$ 0.11  & 0.785 $\pm$ 0.09 & 0.215 $\pm$ 0.10   \\
 \midrule
        Adv 2  & 0.85 $\pm$ 0.32 & 0.15 $\pm$ 0.32 & 0.88 $\pm$ 0.16 & 0.12 $\pm$ 0.16    \\
\bottomrule
\end{tabular}
\label{table:button_status}
\end{center}
\vskip -0.3in
\end{table}

\label{sec:emp_state}

\section{Discussion and Future Work}
The hyperparameter $\delta \in [0,1]$ (Equation \ref{eqn:Loss_Function}) balances the trade-off between the two loss terms. As $\delta$ approaches 1, empathetic state $s_{e}$ becomes similar to original state $s_{i}$, resulting in the inferred rewards of the independent agent being the same as the learning agent's rewards. In practice the bottleneck imposed by the second model ($Q_{learn}$) led to the finding that high $\delta$ values 
%\st{(inducing $s_{e}$ closer to $s_{i}$)} 
were beneficial. In particular, it allowed $s_e$ to reproduce common features such as walls and floors, contributing to better performances. Experimental settings for $\delta$ are in the Supplementary.

The multi-objective nature of the loss term in Equation \ref{eqn:Loss_Function}, can make learning stability hard to achieve (as it is not possible for both loss terms to reach 0 concurrently), and thus could be further improved. Additionally, although there are no limits on the number independent agents modelled, a new Imagination Network would be required for each, which could hinder scalability. Future work can look to address this limitation, perhaps by exploring transfer learning based solutions where multiple imagination transformations are learned with the same network. % Lastly, our assumptions of analogy, full trajectory information and fixed policy for the modelled agent could prevent our work from being applied in certain scenarios.} 
Future work could also look to extend EMOTE to situations where all agents are being trained, and where states are only partially observable. 
%\manisha{\st{and explore ways in which the assumption of analogy could be relaxed.}}\thommen{It is possible to do it if the agents are not analogous in any way?}
%, or are following stochastic policies, 
%\st{which would eschew our current assumption of independent agents operating under fixed policies.}

%to settings where there are at least some state features in the environment which elicit a similar response from the learning agent and an independent. This condition does constrain our work.

\section{Conclusion}
In this work, we presented the EMOTE architecture, which enables a learning agent to model the action-value function of an independent agent, based on its own value function and rewards, under the assumption that agents have ``analagous'' features. A key benefit is the ability to generate an interpretable empathetic states, allowing identification of analogous features between agents. 
%\st{This in turn aids the interpretation of rewards inferred for the independent agent and we contend that it promotes reassurance, helping to reinforce a human's trust in the learning agent.} 
%\thommen{saying that we assume they have analogous goals is more accurate, right? Agents could be analogous in many ways}. 
The architecture is well suited to multiagent learning algorithms, particularly those utilising composite action-value or reward functions in their design. % as our method produces function values at a similar scale between agents. 
%By examining this empathetic state, a human is able to gain insight into the features and goals that are similar or different between the agents, and from this explain the reward values associated for various events. We demonstrated our method by integrating it into the IRL component of a pre-existing Sympathy Framework. 
EMOTE was demonstrated on a previously proposed Sympathy Framework, where it produced more considerate behaviours, more consistent rewards (despite re-configurations of the environment), and also produced insightful empathetic states. We discussed the design of our loss function, and provided insights into future research directions to improve our proposed approach.

%We believe that the proposed approach provides a means of leveraging available information from the learning agent, to inform and assist in modelling another agent. As transparency is a desirable feature of model machine learning methods, the ability to visualise an empathetic state that is interpretable by a human will also be an advantage of such a method.

%\subsubsection{Acknowledgements} Please place your %acknowledgments at
%the end of the paper, preceded by an unnumbered run-in heading %i.e.
%3rd-level heading).

\bibliographystyle{splncs04}
\bibliography{Project2References}
\end{document}